\documentclass{article}
\usepackage{spconf,amsmath,epsfig}

\usepackage{graphicx}
\usepackage{amssymb}

\usepackage[noend]{algpseudocode}
\usepackage{booktabs}
\usepackage{multirow}
\usepackage{graphbox}
\usepackage{array,booktabs}
\usepackage{amsmath,amssymb}
\usepackage{soul}
\usepackage{xcolor}
\usepackage{algorithm,algpseudocode}
\usepackage{url}
\usepackage{subcaption}

\newcommand{\eg}{\emph{e.g.}}
\newcommand{\etal}{\emph{et al.}}
\newcommand{\ie}{\emph{i.e.}}
\newcommand{\etc}{\emph{etc}}
\newcommand{\wrt}{w.r.t. }

\let\OLDthebibliography\thebibliography
\renewcommand\thebibliography[1]{
  \OLDthebibliography{#1}
  \setlength{\parskip}{0pt}
  \setlength{\itemsep}{0pt plus 0.3ex}
}

\begin{document}\sloppy

\def\x{{\mathbf x}}
\def\L{{\cal L}}

\title{MTP: Multi-Task Pruning for Efficient Semantic Segmentation Networks}
\name{Xinghao Chen, Yiman Zhang, Yunhe Wang}
\address{Huawei Noah's Ark Lab\\
\{xinghao.chen, yiman.zhang, yunhe.wang\}@huawei.com}

\maketitle

\begin{abstract}
	This paper focuses on channel pruning for semantic segmentation networks. Previous methods to compress and accelerate deep neural networks in the classification task cannot be straightforwardly applied to the semantic segmentation network that involves an implicit multi-task learning problem via pre-training. To identify the redundancy in segmentation networks, we present a multi-task channel pruning approach. The importance of each convolution filter \wrt the channel of an arbitrary layer will be simultaneously determined by the classification and segmentation tasks. In addition, we develop an alternative scheme for optimizing importance scores of filters in the entire network. Experimental results on several benchmarks illustrate the superiority of the proposed algorithm over the state-of-the-art pruning methods. Notably, we can obtain an about $2\times$ FLOPs reduction on DeepLabv3 with only an about $1\%$ mIoU drop on the PASCAL VOC 2012 dataset and an about $1.3\%$ mIoU drop on Cityscapes dataset, respectively.
\end{abstract}

\section{Introduction}
\label{sec:intro}

In recent years, convolutional neural networks (CNNs) have been the dominant methods for a variety of vision tasks such as image classification, object detection, pose estimation and  segmentation, \etc~\cite{resnet,deeplabv3,chen2020pose,fasterrcnn}. Despite its success, CNN suffers from large model sizes and huge computational resources, making it challenging to be deployed in mobile devices or embedded devices, \eg ~cell phones and cameras.

Various approaches have been proposed to compress and accelerate CNNs, including {channel pruning}~\cite{slimming,he2017channel,thinet,NIPS2018_7367,gao2021network,wang2021convolutional,li2020eagleeye}, quantization~\cite{zhou2016dorefa}, distillation~\cite{chen2020optical} and lightweight network design. Among them channel pruning is one of the most popular methods to accelerate over-parameterized CNNs, since the pruned deep networks can be directly applied on any off-the-shelf platforms and hardware to obtain the online speed-up. However, most of existing channel pruning methods are dedicated to image classification task on a particular dataset, \eg ThiNet~\cite{thinet}, network slimming~\cite{slimming}. 

\begin{figure*}[t]
	\footnotesize
	\begin{center}
		\begin{tabular}{c}
			\includegraphics[width = 0.88\linewidth]{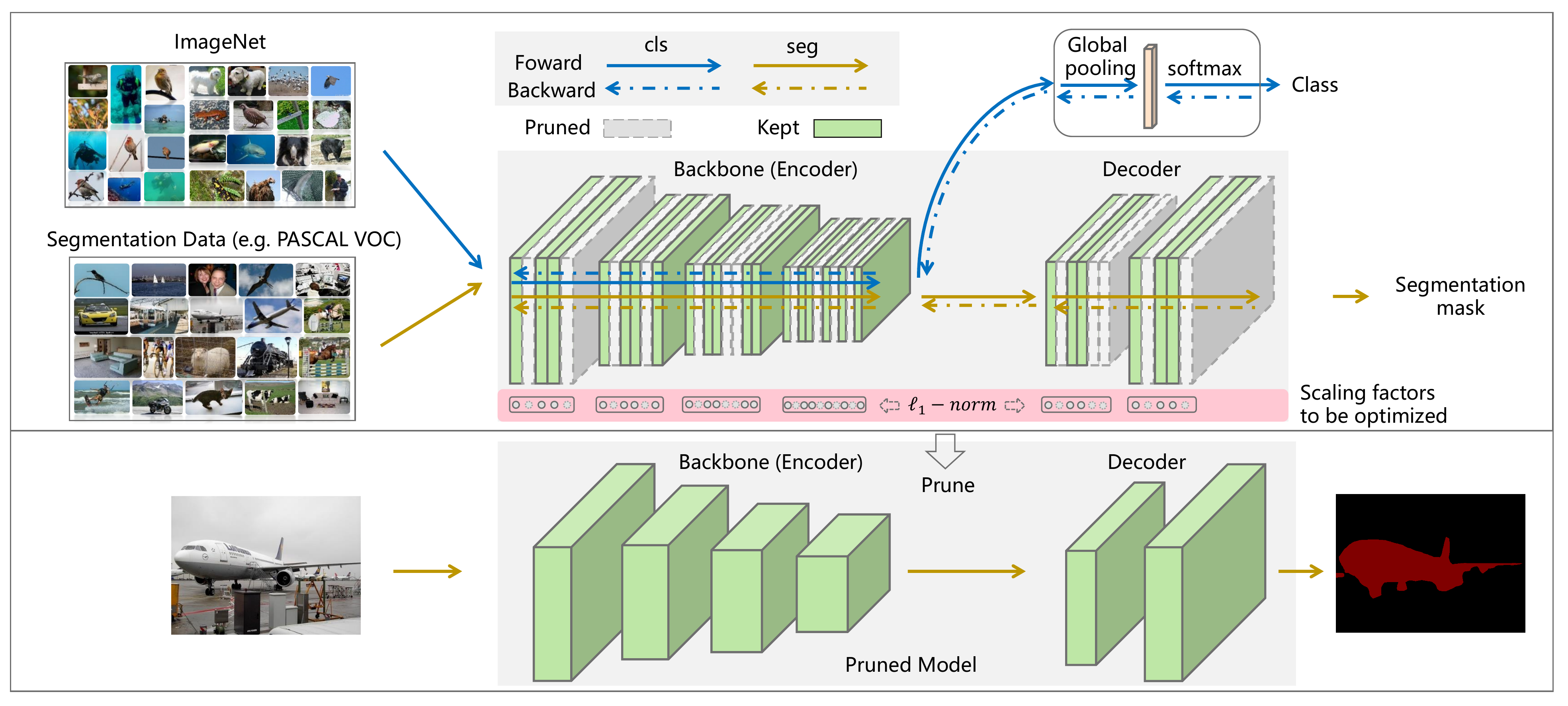} \\
		\end{tabular}
	\end{center}
	\vspace{-2em}
	\caption{%
		\small The diagram of the proposed multi-task pruning for semantic segmentation networks. The scaling factor of each convolution filter in the given network will be simultaneously determined by the classification and segmentation tasks.
	}
	\label{fig:framework}
	\vspace{-1em}
\end{figure*}

Efficient semantic segmentation networks are of great importance for deployment on mobile devices. Nevertheless, there are only few works discussing the channel pruning for the semantic segmentation neural networks. Luo~\etal~\cite{thinet} pruned filters in the backbone network on ImageNet and transferred it to the segmentation network. Besides, pruning methods designed over the classification task have been straightforwardly applied to segmentation neural networks~\cite{huang2018learning,pcas,He_2021_WACV}, but they rarely have a thorough analysis on the influence of ImageNet pre-training on compressing segmentation neural networks. 

\begin{table}
	\begin{center}
		\scalebox{0.9}{
		\begin{tabular}{l|c}
			\toprule
			Methods & mIoU (\%) \\ 
			\midrule
			DeepLabV3~\cite{deeplabv3} &  77.27\\ 
			Slimming 0.5$\times$ w/o ImageNet finetuning &  70.81\\
			Slimming 0.5$\times$ w/ ImageNet finetuning & 74.91\\
			\bottomrule
		\end{tabular}
		}
		\caption{\small The impact of ImageNet fine-tuning for pruned segmentation networks on PASCAL VOC 2012.}
		\label{tab:pretrain}
		\vspace{-2.5em}
	\end{center}
\end{table}

The state-of-the-art semantic segmentation methods follow the \textit{de facto} paradigm to pre-train models on a large-scale classification dataset (\textit{e.g.} ImageNet) and then fine-tune models on target segmentation datasets to achieve a satisfactory segmentation performance. The main underlying motivation is to transfer latent knowledge learned from the large-scale classification dataset to facilitate the training of semantic segmentation. Recently some efforts have been devoted to rethink the necessity of ImageNet pre-training and train an object detector or semantic segmentation network from scratch~\cite{he2018rethinking,autodeeplab}. However, ImageNet pre-training still plays \textcolor{black}{a} fatal role in current state-of-the-art semantic segmentation networks.
In particular, if we prune some unimportant neurons in the DeepLabv3~\cite{deeplabv3} and only fine-tune it on the target segmentation dataset, the accuracy drop is 6.46\%. This drop can be further reduced to 2.36\% by fine-tuning the backbone part on the ImageNet dataset as shown in Table~\ref{tab:pretrain}.

Beyond the image classification, this paper focuses on the compression problem of semantic segmentation networks. The pruned architecture should be simultaneously determined by the classification and segmentation tasks, since the latent knowledge learned from the large-scale classification dataset is crucial for semantic segmentation as illustrated in Table~\ref{tab:pretrain}.
Specifically, each convolution filter in the given deep network will be assigned with an importance score and these scores will be optimized on both classification and segmentation datasets. An $\ell_1$-norm is introduced to shrink importance score vectors and thus redundant filters will be removed. In addition, we develop an effective alternative optimization scheme to solve the multi-task pruning problem as \textcolor{black}{shown} in Figure~\ref{fig:framework}. 
Extensive experiments conducted on several benchmark datasets demonstrate the significance of pruning segmentation networks through a multi-task scheme and the advantages of the proposed algorithm over existing methods.

\section{Multi-task Pruning}
\label{sec:mtp}

Generally, most of existing models for visual segmentation~\cite{deeplabv3plus,panet}) contain two training stages, \ie ~pre-training and fine-tuning. In practice, the backbone part is usually pre-trained on the ImageNet dataset for having a better initialization, \ie
\begin{equation}
	\small\scalebox{0.85}{$ \displaystyle
	\tilde{
		\mathcal{W}}_{1} = \mathop{\arg\min}_{\mathcal{W}_1} \ell_{cls}(\mathcal{N}_1(\mathcal{W}_1,\mathcal{X}_1), \mathcal{Y}_{1}),$}
	\label{Fcn:class}
\end{equation}
where $\mathcal{N}_1$ is the backbone network, $\mathcal{X}_1$ is the classification training sample and $\mathcal{Y}_1$ is the corresponding ground-truth label, $\mathcal{W}_1$ denotes the parameters in $\mathcal{N}_1$, and $\ell_{cls} (\cdot)$ is the classification loss function. 

Then, both the backbone and the decoder part (\eg~ atrous spatial pyramid pooling, ASPP in DeepLabv3~\cite{deeplabv3}) are further optimized on the subsequent \textcolor{black}{semantic} segmentation task as follows:
\begin{equation}
	\small\scalebox{0.85}{$ \displaystyle
	\mathcal{W}_1^*, \mathcal{W}_2^* = \mathop{\arg\min}_{\mathcal{W}_1,\mathcal{W}_2} \ell_{seg}(\mathcal{N}_2(\mathcal{W}_1,\mathcal{W}_2,\mathcal{X}_2), \mathcal{Y}_2),$}
	\label{Fcn:seg}
\end{equation}
where $\mathcal{N}_2$ is the whole \textcolor{black}{semantic} segmentation network including the backbone and decoder, $\mathcal{W}_2$ denotes the parameters for decoder part, $\mathcal{X}_2$ and $\mathcal{Y}_2$ are the training images and their corresponding ground-truth labels in the \textcolor{black}{semantic} segmentation task, respectively. $\mathcal{W}_1$ is initialized with $\tilde{\mathcal{W}}_1$ solved from Eq.~(\ref{Fcn:class}), and $\ell_{seg} (\cdot)$ is the loss function in the segmentation task. Note that decoder-encoder structure is widely used in \textcolor{black}{semantic} segmentation network~\cite{deeplabv3,deeplabv3plus,pspnet}. For DeepLabv3~\cite{deeplabv3}, ASPP can be viewed as a vanilla decoder.

Neural architectures are usually designed with a large volume of neurons and parameters for better accuracy, which makes the deployment of these networks on mobile devices very difficult. Although there are a number of approaches proposed for eliminating redundancy in deep neural networks, most of them are explored for processing networks for visual classification~\cite{thinet,slimming}. Generally, the pruning for the classification network \wrt $\mathcal{N}_1$ can be formulated as:
\begin{equation}
	\small\scalebox{0.85}{$ \displaystyle
	\min_{\mathcal{W}_1} \ell_{cls}(\mathcal{N}_1(\mathcal{W}_1,\mathcal{X}_1), \mathcal{Y}_{1})+\alpha_1||\gamma_1||_1,$}
	\label{Fcn:classprun}
\end{equation}
where $||\cdot||_1$ is the $\ell_1$-norm for making resulting network sparse, $\alpha_1$ is the hyper-parameter for controlling the sparsity of network $\mathcal{N}_1$. $\gamma_1$ are scaling factors assigned to all channels to indicate their importance scores. In practice, the scaling parameters in batch normalization layers are leveraged as the channel scaling factors $\gamma_1$, following good practice in network slimming~\cite{slimming,ye2018rethinking}. Therefore, $\gamma_1$ is part of the network parameters $\mathcal{W}_1$. In other words, we impose $\ell_1$-norm on part of network parameters $\mathcal{W}_1$ so that we can obtain a model with sparse scaling factors for channels. If the scaling factor for a channel is near zero, then we can eliminate this channel without much impact on the final performance.

Similarly, we can also directly utilize the above function to remove redundant parameters in the network $\mathcal{N}_2$, in other words, applying traditional pruning methods (\eg \cite{thinet,slimming}) on segmentation networks. However, some experimental evidences shown in Table~\ref{tab:pretrain} illustrate that the pruned backbone $\mathcal{N}_1$ should be further tuned on the ImageNet dataset for maintaining the accuracy. This observation motivates us to design a new pruning scheme that simultaneously determines the pruned model via a multi-task method on the tasks of ImageNet classification and semantic segmentation. In this case, we may be able to obtain a more optimal pruned architecture for segmentation. Specifically, we want to optimize the following formula to obtain a sparse model:
\begin{equation}
	\small\scalebox{0.85}{$ \displaystyle
	\begin{aligned}
		\min_{\mathcal{W}_1,\mathcal{W}_2} &\ell_{cls}(\mathcal{N}_1(\mathcal{W}_1,\mathcal{X}_1), \mathcal{Y}_{1})
		+ \lambda \ell_{seg}(\mathcal{N}_2(\mathcal{W}_1, \mathcal{W}_2,\mathcal{X}_2), \mathcal{Y}_{2})\\
		&+ \alpha_1||\gamma_1||_1+\alpha_2||\gamma_2||_1, 
		\label{Fcn:prun}
	\end{aligned} $}
\end{equation}
where $\lambda$ is the hyper-parameter for seeking the trade-off between the classification and the segmentation tasks, $\alpha_2$ is the weight parameter for the sparsity of $\mathcal{W}_2$ and $\gamma_2$ are scaling factors for channels in decoder. After minimizing Eq.~(\ref{Fcn:prun}), we can obtain a sparse network and prune the unimportant channels for a lightweight semantic segmentation network.

\section{Optimization}
\label{sec:opt}

We have proposed a multi-task pruning approach to produce lightweight semantic segmentation networks. In this section, we proceed to introduce an optimization method to solve the proposed multi-task pruning problem.
Eq.~(\ref{Fcn:prun}) involves two sub-networks and two different tasks and datasets, which is hard to directly optimize. We first add an auxiliary variable:
\begin{equation}
	\small\scalebox{0.8}{$ \displaystyle
	\begin{aligned}
		\min_{\mathcal{W}_1,\mathcal{W}_2,\mathcal{W}_3}& \ell_{cls}(\mathcal{N}_1(\mathcal{W}_1,\mathcal{X}_1), \mathcal{Y}_{1}) + \lambda\ell_{seg}(\mathcal{N}_2(\mathcal{W}_3, \mathcal{W}_2,\mathcal{X}_2), \mathcal{Y}_{2}) \\
		& + \alpha_1||\gamma_1||_1+\alpha_2(||\gamma_2||_1+||\gamma_3||_1), \;\;s.t.\;\;\mathcal{W}_1=\mathcal{W}_3,
	\end{aligned} $}
	\label{Fcn:relax}
\end{equation}
where $\gamma_3$ are scaling factors and are part of $\mathcal{W}_3$.
By introducing $\mathcal{W}_3$, the above function can now be easily solved by exploiting the inexact augmented Lagrange multiplier. We further introduce multipliers $\mu$ and $E$, the loss function of Eq.~(\ref{Fcn:relax}) can be formulated as
\begin{equation}
	\small\scalebox{0.75}{$ \displaystyle
	\begin{aligned}
		&\mathcal{L}(\mathcal{W}_1,\mathcal{W}_2,\mathcal{W}_3,\mu,E) = \ell_{cls}(\mathcal{N}_1(\mathcal{W}_1,\mathcal{X}_1), \mathcal{Y}_{1})+
		\lambda\ell_{seg}(\mathcal{N}_2(\mathcal{W}_3, \mathcal{W}_2,\mathcal{X}_2), \mathcal{Y}_{2})\\&+\frac{\mu}{2}||\mathcal{W}_1-\mathcal{W}_3||_2^2
		+\langle E,\mathcal{W}_1-\mathcal{W}_3 \rangle+\alpha_1||\gamma_1||_1+\alpha_2(||\gamma_2||_1+||\gamma_3||_1).
	\end{aligned} $}
	\label{Fcn:lag}
\end{equation}

Then, the optimal weights of the desired sparse segmentation network can be obtained by updating $\mathcal{W}_1$, $\mathcal{W}_2$, $\mathcal{W}_3$, alternately.

\noindent\textbf{Solve $\mathcal{W}_1$:} The loss function for optimizing $\mathcal{W}_1$ is 
\begin{equation}
	\small\scalebox{0.8}{$ \displaystyle
	\begin{aligned}
		\mathcal{L}_1(\mathcal{W}_1,\mu,E) = &\ell_{cls}(\mathcal{N}_1(\mathcal{W}_1,\mathcal{X}_1), \mathcal{Y}_{1})
		+\langle E,\mathcal{W}_1-\mathcal{W}_3 \rangle\\
		&+\frac{\mu}{2}||\mathcal{W}_1-\mathcal{W}_3||_2^2+\alpha_1||\gamma_1||_1,
	\end{aligned}$}
	\label{Fcn:Lw1}
\end{equation}
which can be optimized using the backbone network $\mathcal{N}_1$ on the classification dataset $\mathcal{X}_1$, $\mathcal{Y}_1$.

\noindent\textbf{Solve $\mathcal{W}_2$:} The loss function of the weight in the decoder network can be written as
\begin{equation}
	\small\scalebox{0.8}{$ \displaystyle
	\begin{aligned}
		\mathcal{L}_2(\mathcal{W}_2) = \lambda\ell_{seg}(\mathcal{N}_2(\mathcal{W}_3, \mathcal{W}_2,\mathcal{X}_2), \mathcal{Y}_{2})+\alpha_2||\gamma_2||_1.
	\end{aligned}$}
	\label{Fcn:Lw2}
\end{equation}
\textbf{Solve $\mathcal{W}_3$:} The loss \wrt the auxiliary variable $\mathcal{W}_3$ is
\begin{equation}
	\small\scalebox{0.8}{$ \displaystyle
	\begin{aligned}
		\mathcal{L}_3(\mathcal{W}_3,\mu,E) = &\lambda\ell_{seg}(\mathcal{N}_2(\mathcal{W}_3, \mathcal{W}_2,\mathcal{X}_2), \mathcal{Y}_{2})
		+\frac{\mu}{2}||\mathcal{W}_1-\mathcal{W}_3||_2^2+ \\
		&\langle E,\mathcal{W}_1-\mathcal{W}_3 \rangle + \alpha_2||\gamma_3||_1,
	\end{aligned}$}
	\label{Fcn:Lw3}
\end{equation}
which can be only optimized on the segmentation dataset.

In addition, the multipliers are updated as: 
\begin{equation}
	\small\scalebox{0.8}{$ \displaystyle
	E = E+\mu(\mathcal{W}_1-\mathcal{W}_3),\quad \mu = \rho \mu, $}
	\label{Fcn:mul}
\end{equation}
where $\rho>1$ is a constant.

\begin{algorithm}[htb]
	\small
	\caption{Multi-task Pruning (MTP).}
	\label{Alg:main}
	\begin{algorithmic}[1]
		\Require A segmentation network including the backbone part $\mathcal{N}_1$ and the decoder part $\mathcal{N}_2$ and their initial weights $\mathcal{W}_1$ and $\mathcal{W}_2$. Training datasets and ground-truth $\mathcal{X}_1$, $\mathcal{X}_2$, $\mathcal{Y}_1$, and $\mathcal{Y}_2$, respectively, parameters $\lambda$, $\alpha_1$, $\alpha_2$, $\rho$.
		\State \textbf{Training a sparse model:}
		\Repeat
		\State Optimize $\mathcal{W}_1$ according to Eq.~\ref{Fcn:Lw1}.
		\State Optimize $\mathcal{W}_2$ according to Eq.~\ref{Fcn:Lw2}.
		\State Optimize $\mathcal{W}_3$ according to Eq.~\ref{Fcn:Lw3}.
		\State Update $E$ and $\mu$ according to Eq.~\ref{Fcn:mul}.
		\Until convergence
		\State \textbf{Pruning and Fine-tuning:}
		\State Calculate scaling factors $\gamma_1$ and $\gamma_2$ from the parameters $\mathcal{W}_1$ and $\mathcal{W}_2$ of sparse model.
		\State Prune the model according to $\gamma_1$ and $\gamma_2$.
		\State Fine-tune the pruned model $\tilde{\mathcal{N}}$ on ImageNet and target segmentation dataset sequentially.
		\Ensure The pruned model $\tilde{\mathcal{N}}$.
	\end{algorithmic}	
\end{algorithm}
\setlength{\textfloatsep}{5pt}%

By optimizing the Eq.~(\ref{Fcn:Lw1}) - Eq.~(\ref{Fcn:mul}), we can obtain a sparse model that has small scaling factors in some channels. Eliminating channels with near-zero scaling factors results in a pruned and lightweight segmentation network. Given a  predefined global percentile, the threshold of scaling factor values is calculated and all channels with scaling factors below the threshold will be pruned. Since the scaling factors in backbone and decoder network are optimized alternately, setting the same global threshold for both the backbone and decoder is inappropriate. Therefore, we instead use two independent thresholds for these two parts of the segmentation network. Specifically, if we want to prune a certain percentile (\textcolor{black}{denoted as} $p\%$, \ie $p=50$) of all channels, we set the threshold for the backbone (\textcolor{black}{denoted as} $\tau_1$) so that $p\%$ of all channels in backbone have smaller scaling factors than $\tau_1$. Similarly, the threshold for decoder $\tau_2$ is set to eliminate $p\%$ of channels in decoder network.
To further recover the performance drop of the pruned model, we fine-tune the models for a few epochs on ImageNet and segmentation dataset. The proposed multi-task pruning procedure for semantic segmentation network is summarized as shown in Algorithm~\ref{Alg:main}.

\section{Experiments}
\label{sec:exp}

\begin{table*}[tb]
	\centering
	\small
	\setlength{\tabcolsep}{5.5mm}
	\scalebox{0.83}{
	\begin{tabular}{ll|llll}
		\toprule
		& & mIoU (\%) & \#Params (M) & FLOPs (G)\textsuperscript\textdagger & GPU Speed (ms)\textsuperscript\textdagger \\ 
		\midrule
		& DeepLabV3-R101~\cite{deeplabv3} & 77.27 & 58.04  & 71.52 & 39.67  \\ 
		\midrule
		\multirow{3}{*}{0.75$\times$} & Uniform  & 75.09 $_{\downarrow 2.18}$ & 40.18 $_{0.69\times}$ & 49.70 $_{0.69\times}$ & 33.83 $_{0.85\times}$\\ 
		&Slimming~\cite{slimming} & 76.64 $_{\downarrow 0.63}$ & 43.11 $_{0.74\times}$ & 52.93 $_{0.74\times}$ & 34.93 $_{0.88\times}$ \\ 
		&\bf MTP (Ours) & \bf 77.28 $_{\bf \uparrow 0.01}$ & 44.32 $_{0.76\times}$ & 54.89 $_{0.77\times}$ & 34.86 $_{0.88\times}$ \\ 
		\midrule
		\multirow{3}{*}{0.5$\times$} & ThiNet~\cite{thinet} & 74.71 $_{\downarrow 2.56}$ & 32.88 $_{0.57\times}$ & 39.55 $_{0.55\times}$ & 30.51 $_{0.78\times}$ \\ 		
		&Slimming~\cite{slimming} & 74.91 $_{\downarrow 2.36}$ & 28.61 $_{0.49\times}$ & 35.96 $_{0.50\times}$ & 31.52 $_{0.79\times}$ \\ 
		&\bf MTP (Ours) & \bf 76.29 $_{\bf \downarrow 0.98}$ & 30.33 $_{0.52\times}$ & 38.87 $_{0.54\times}$ & 31.30 $_{0.79\times}$ \\ 
		\midrule
		&PSPNet-R50~\cite{pspnet} & 77.05 & 46.71 & 190.43  & 73.62\\ 
		&PSANet-R50~\cite{psanet} & 77.25 & 50.81 & 205.98& 62.09\\ 
		\bottomrule
	\end{tabular}
	}
	\vspace{-0.8em}
	\caption{\small Comparisons with {state-of-the-art} pruning methods on Pascal VOC 2012 \textit{val} set. OS: output stride. \textsuperscript\textdagger Image size $513 \times 513$.}
	\label{tab:voc}
	\vspace{-0.7em}
\end{table*}

\begin{figure*}%
	\centering
	\begin{minipage}{0.6\linewidth}
		\centering
		\small\scalebox{0.83}{
			\begin{tabular}{ll|llll}
				\toprule
				&& mIoU (\%) & \#Params (M) & FLOPs (G)\textsuperscript\textdagger & GPU Speed (ms)\textsuperscript\textdagger \\ 
				\midrule
				&{DeepLabV3-R101~\cite{deeplabv3}}     & 78.65 & 58.04  & 201.88 & 183.07  \\ 
				\midrule
				\multirow{2}{*}{0.75$\times$} & Slimming~\cite{slimming} & 78.37 $_{\downarrow 0.28}$ & 44.43 $_{0.77\times}$ & 155.08 $_{0.77\times}$ & 164.58 $_{0.86\times}$ \\ 
				&\bf MTP (Ours) & \bf 78.60 $_{\bf \downarrow 0.05}$ & 45.08 $_{0.78\times}$ & 158.10 $_{0.78\times}$ & 164.36 $_{0.86\times}$ \\ 
				\midrule
				\multirow{2}{*}{0.5$\times$} & Slimming~\cite{slimming} & 76.94 $_{\downarrow 1.71}$ & 29.74 $_{0.51\times}$ & 106.02 $_{0.53\times}$ & 132.98 $_{0.67\times}$ \\ 
				&\bf MTP (Ours) & \bf 77.39 $_{\bf \downarrow 1.26}$ & 31.16 $_{0.56\times}$ & 112.52 $_{0.54\times}$ & 128.71 $_{0.69\times}$ \\ %
				\midrule\midrule
				&{BiSeNet-R18~\cite{bisenet}} &74.83 & 12.89 & 104.27 & 29.11\\ 
				\midrule
				\multirow{3}{*}{0.75$\times$} & Slimming~\cite{slimming} & 72.99 $_{\downarrow 1.84}$ & 9.23 $_{0.72\times}$& 86.22 $_{0.83\times}$ & 25.58 $_{0.88\times}$ \\ 
				&FPGM~\cite{he2019filter} &73.10 $_{\downarrow 1.73}$&9.14 $_{0.71\times}$&87.30 $_{0.84\times}$&25.89 $_{0.89\times}$\\
				&\bf MTP (Ours) & \bf 73.46 $_{\bf \downarrow 1.37}$ &9.22 $_{0.72\times}$ & 88.33 $_{0.85\times}$ & 26.02 $_{0.89\times}$\\ 
				\midrule
				\multirow{3}{*}{0.5$\times$} & Slimming~\cite{slimming} & 71.81 $_{\downarrow 3.02}$ & 5.94 $_{0.46\times}$ & 71.65 $_{0.69\times}$ & 23.48 $_{0.80\times}$ \\ 
				&FPGM~\cite{he2019filter} &71.98 $_{\downarrow 2.85}$&5.90 $_{0.46\times}$&73.65 $_{0.71\times}$&23.61 $_{0.81\times}$\\
				&\bf MTP (Ours) &\bf 72.45 $_{\bf \downarrow 2.38}$ &5.78 $_{0.45\times}$ & 75.48 $_{0.72\times}$ & 23.80 $_{0.82\times}$\\ 
				\bottomrule
		\end{tabular} }
		\vspace{-0.4em}
		\captionof{table}{\small Comparisons with state-of-the-arts on Cityscapes \textit{val} set. All models are trained on \textit{train\_fine} set without pretraining on COCO. OS: output stride. \textsuperscript\textdagger Image size $2048 \times 1024$.}
		\label{tab:cityscapes}
	\end{minipage}
	\hfill
	\begin{minipage}{0.38\linewidth}
		\centering
		\begin{subfigure}[t]{.79\linewidth}
			\includegraphics[width=\textwidth]{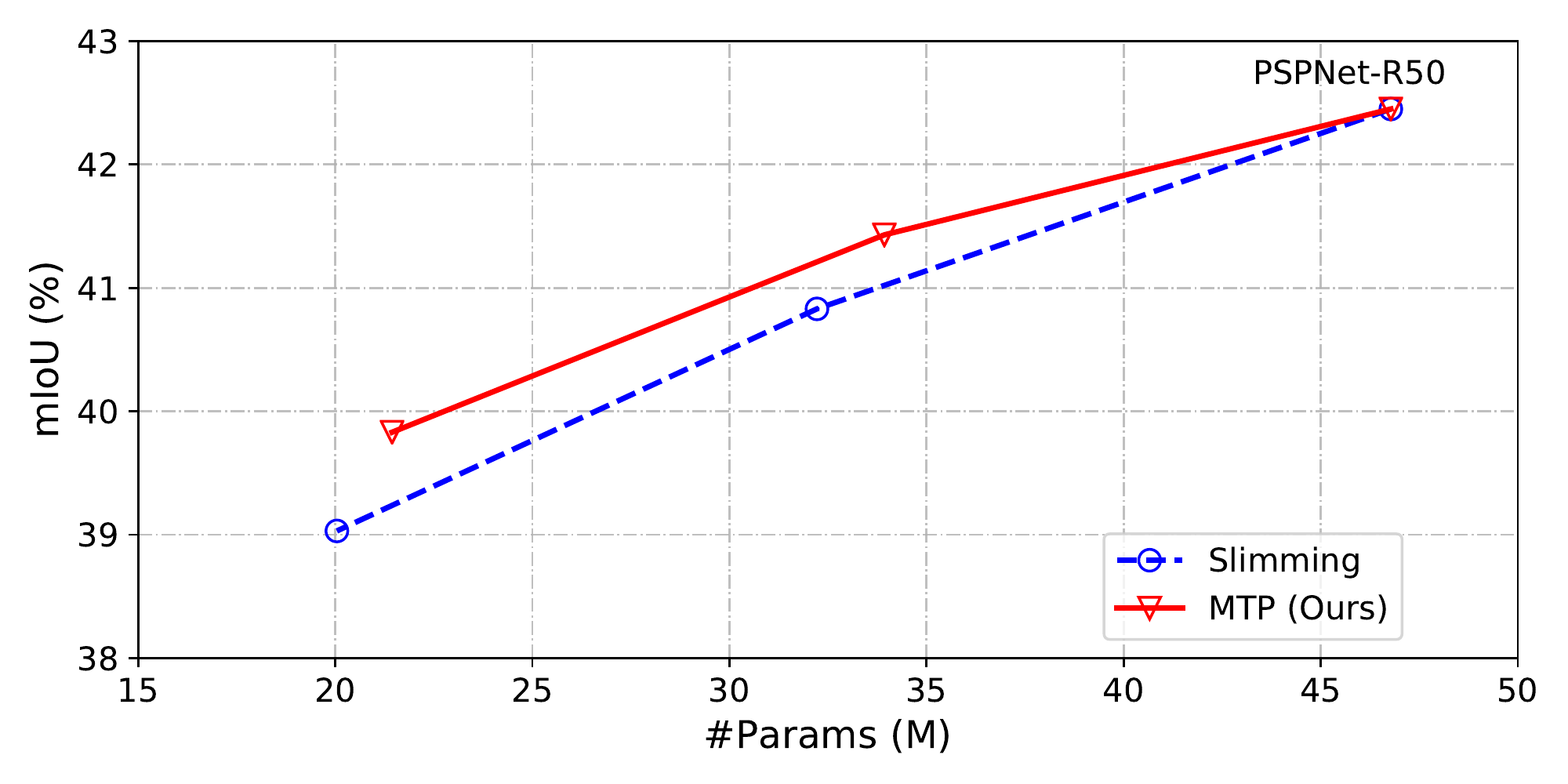}
			\vspace{-1.7em}
			\captionof{figure}{\small\small Comparisons with prior method.}
			\label{fig:ade20k_pruning}
		\end{subfigure} \\
		\begin{subfigure}[b]{.8\linewidth}
			\includegraphics[width=\textwidth]{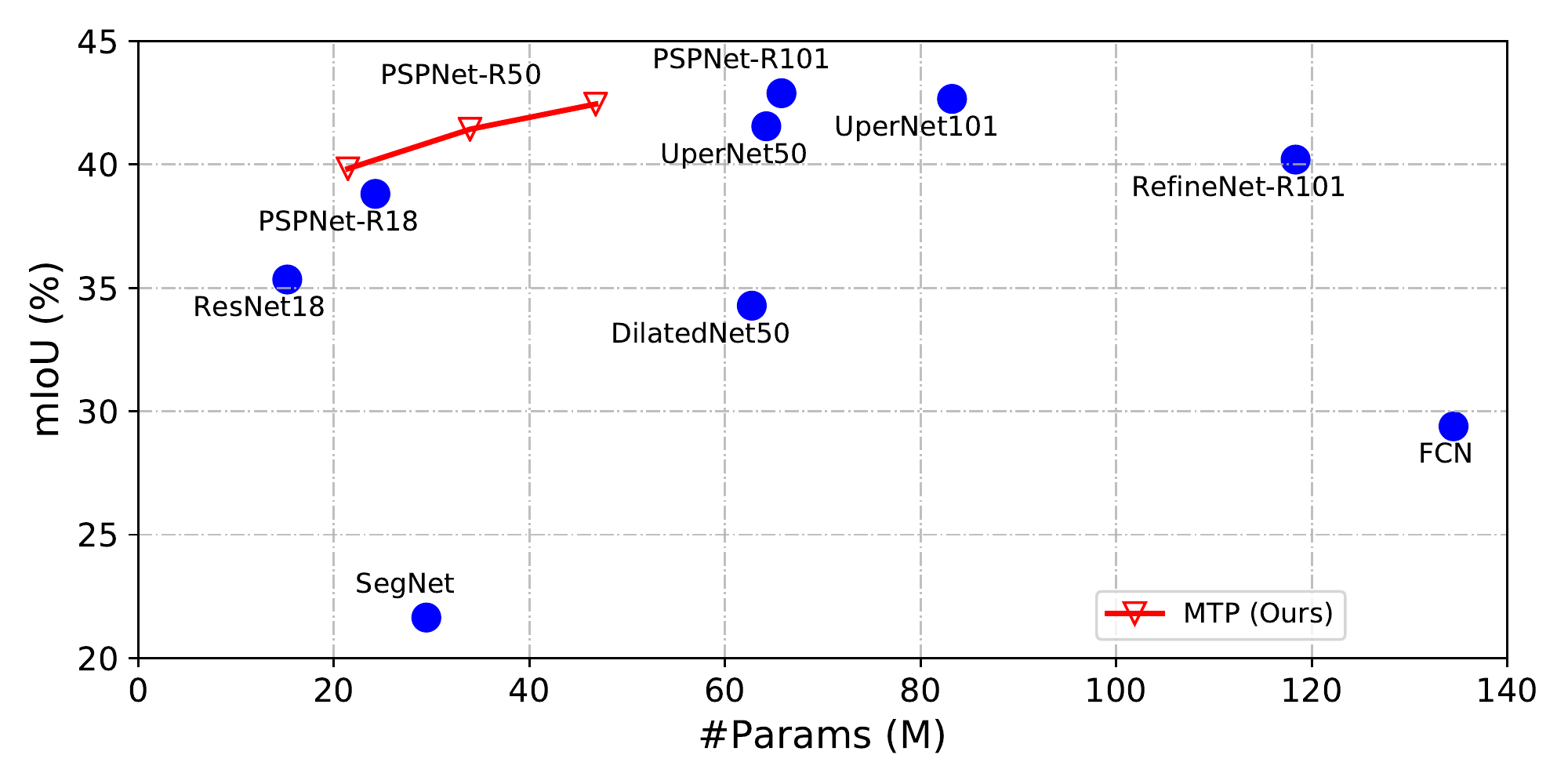}
			\vspace{-1.7em}
			\captionof{figure}{\small Comparisons with state of the arts.}
			\label{fig:ade20k}
		\end{subfigure} 
		\vspace{-0.9em}
		\captionof{figure}{Results on ADE20K \textit{val} set.}
	\end{minipage}
	\vspace{-1.2em}
\end{figure*}

In this section we first conduct experiments on several challenging benchmarks (including PASCAL VOC 2012~\cite{pascal}, Cityscapes~\cite{cityscapes} and ADE20K~\cite{zhou2017scene}). We apply the proposed multi-task pruning method on several competitive semantic segmentation models, including DeepLabv3~\cite{deeplabv3}, PSPNet~\cite{pspnet} and a real-time model BiSeNet~\cite{bisenet}. We then further introduce more ablation analysis on the proposed method.%

\subsection{Pruning DeepLabv3 on PASCAL VOC 2012}

We choose DeepLabv3~\cite{deeplabv3}\footnote{We use the implementation of DeepLabv3 at \url{https://github.com/chenxi116/DeepLabv3.pytorch}.} as our baseline. We use ResNet-101 as the backbone of DeepLabv3 with multi\_grid=(1,1,1) and the output stride is 16. 

When pruning 25\% of channels, our proposed method (MTP 0.75$\times$) suffers no performance drop while reducing the number of parameter and FLOPs to 77\%, as shown in Table~\ref{tab:voc}. In contrast, the model whose number of channels in each layer are uniformly set to 75\% of original model (denoted as Uniform 0.75$\times$) achieves only 75.09\% mIoU, which is 2.19\% worse than our proposed method. 

We also compare our method with some prior pruning methods, including ThiNet~\cite{thinet} and network slimming~\cite{slimming}. These pruning methods are originally proposed for classification task. We adapt these methods for pruning semantic segmentation networks. For example, we apply the method of network slimming to DeepLabv3, \ie,~add sparse constraints to the scaling factors of each channel and train the DeepLabv3 on PASCAL VOC 2012. Then we eliminate channels with small scaling factors to prune the model and fine-tune it on ImageNet and PASCAL VOC 2012 sequentially. \textcolor{black}{Network slimming is actually the single-task pruning method. Our MTP-0.5$\times$ achieves 1.38\% higher mIoU than Slimming-0.5$\times$, which justify the advantage of our MTP over single task pruning.} Meanwhile, ThiNet achieves the performance of 74.71\% when pruning 50\% channels and is outperformed by our method with a margin of 1.66\%. Our method reduces the number of parameters and FLOPs to about 50\% when suffering from only less than 1\% mIoU drop. Most importantly, our pruned model achieves actual speedup and reduce the inference time on GPU to 79\%. 

We also compare our pruned models with some state-of-the-art segmentation networks, including PSPNet-R50~\cite{pspnet} and PSANet-R50~\cite{psanet}\footnote{\url{https://github.com/hszhao/semseg}}, as shown in Table~\ref{tab:voc}. The pruned model obtained by our method achieves better mIoU than PSPNet-R50 while having significant fewer FLOPs and higher speed. When compared with PSANet-R50, our model achieves similar mIoU and runs about $1.8 \times$ faster.

\subsection{Pruning DeepLabv3 on Cityscapes}

As shown in Table~\ref{tab:cityscapes}, network slimming~\cite{slimming} has mIoU of 78.37\% and suffers from performance drop of 0.28\% when pruning 25\% of all channels. In contrast, our proposed method boosts the performance of pruned model and has insignificant accuracy drop. When the pruning rate is higher, the superiority of our method over state-of-the-art pruning methods becomes more significant. Specifically, when the model size and FLOPs are reduced to about 50\%, our pruned model only has 1.26\% performance drop, 0.45\% less than network slimming~\cite{slimming}. MTP 0.5$\times$ achieves mIoU of 77.39\% and reduces the latency to only 69\% of original one.

We also evaluate our method on Cityscapes \textit{test} set. The baseline DeepLabv3 achieves mIoU of 78.42\% and MTP 0.5$\times$ obtains mIoU of 77.41\%, suffering from accuracy drop of only 1\% while having around 2$\times$ reduction of FLOPs.

\subsection{Pruning BiSeNet on Cityscapes}
We further conduct experiments for pruning on lightweight semantic segmentation networks, \eg~ BiSeNet~\cite{bisenet}, to demonstrate the scalability of our proposed method. Note that pruning a lightweight model is much more challenging since it contains less redundancy. As shown in Table~\ref{tab:cityscapes}, the baseline BiSeNet with ResNet-18 achieves the mIoU of 74.83\% on Cityscapes \textit{val} set with 104.27 GFLOPs. Our method obtains mIoU of 73.46\% when keeping 75\% channels, which leads to 1.37\% mIoU drop. Nevertheless, our method still outperforms prior pruning methods, \eg network slimming~\cite{slimming} and FPGM~\cite{he2019filter} by 0.47\% and 0.36\%, respectively. Moreover, MTP-0.5$\times$ achieves 28\% FLOPs reduction with 2.38\% mIoU drop and consistently performs better than prior pruning methods. We also evaluate the actual inference speed of our pruned models on GPU. MTP-0.5$\times$ reduces the GPU latency from 29.11 ms to 23.80 ms, which achieves better accuracy than prior pruning methods with similar latency.

\subsection{Pruning PSPNet on ADE20K}
Here we explore the performance of our proposed method on a more challenging scene parsing benchmark, \ie ADE20K~\cite{zhou2017scene}.
As shown in Figure~\ref{fig:ade20k_pruning}, our method suffers only 0.58 mIoU drop when pruning 25\% channels, which is 0.24\% better than network slimming~\cite{slimming}. Pruning more channels leads to slightly worse mIoU but consistently outperforms prior method.
We also compare our pruned models with state-of-the-art semantic segmentation models, as shown in Figure~\ref{fig:ade20k}. Our models (MTP-0.75$\times$ and MTP-0.5$\times$) achieve a better trade-off between accuracy and model size. Specifically, MTP-0.5$\times$ has higher mIoU and fewer parameters than PSPNet-R18~\cite{pspnet}. MTP-0.75$\times$ obtains similar accuracy with UperNet50~\cite{xiao2018unified} and RefineNet-R101~\cite{lin2017refinenet} but with 47\% and 71\% fewer parameters, respectively.

\begin{figure}[tb]
	\begin{minipage}[t]{0.95\linewidth}
		\centering
		\includegraphics[width = 0.8\linewidth]{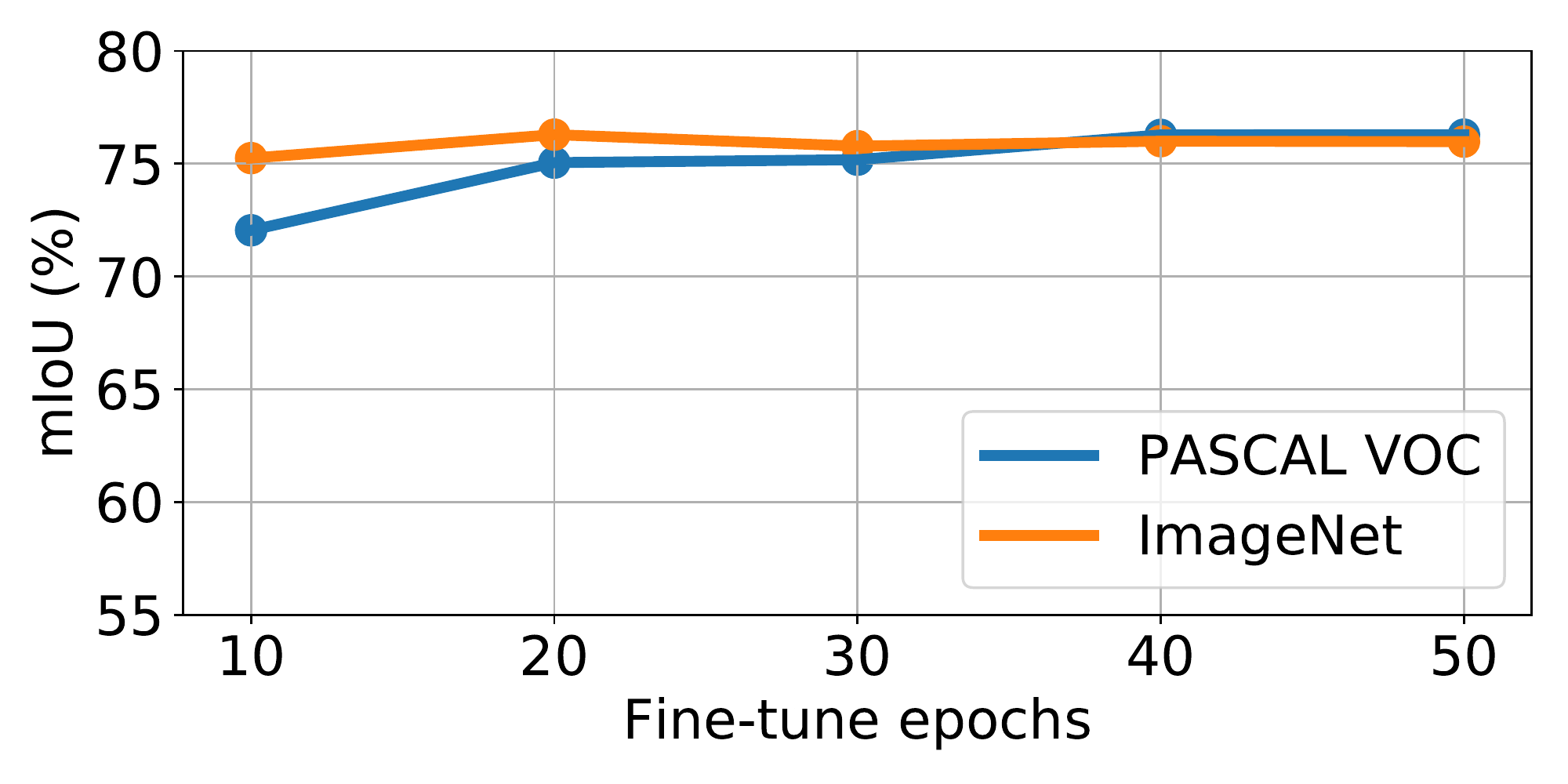}
		\vspace{-1em}
		\caption{\small Impact of fine-tuning epochs on ImageNet and VOC.}
		\label{fig:fine-tune-epochs}
	\end{minipage}
	\hspace{1.ex}
	\begin{minipage}[t]{0.95\linewidth}
		\centering
		\includegraphics[width = 0.8\linewidth]{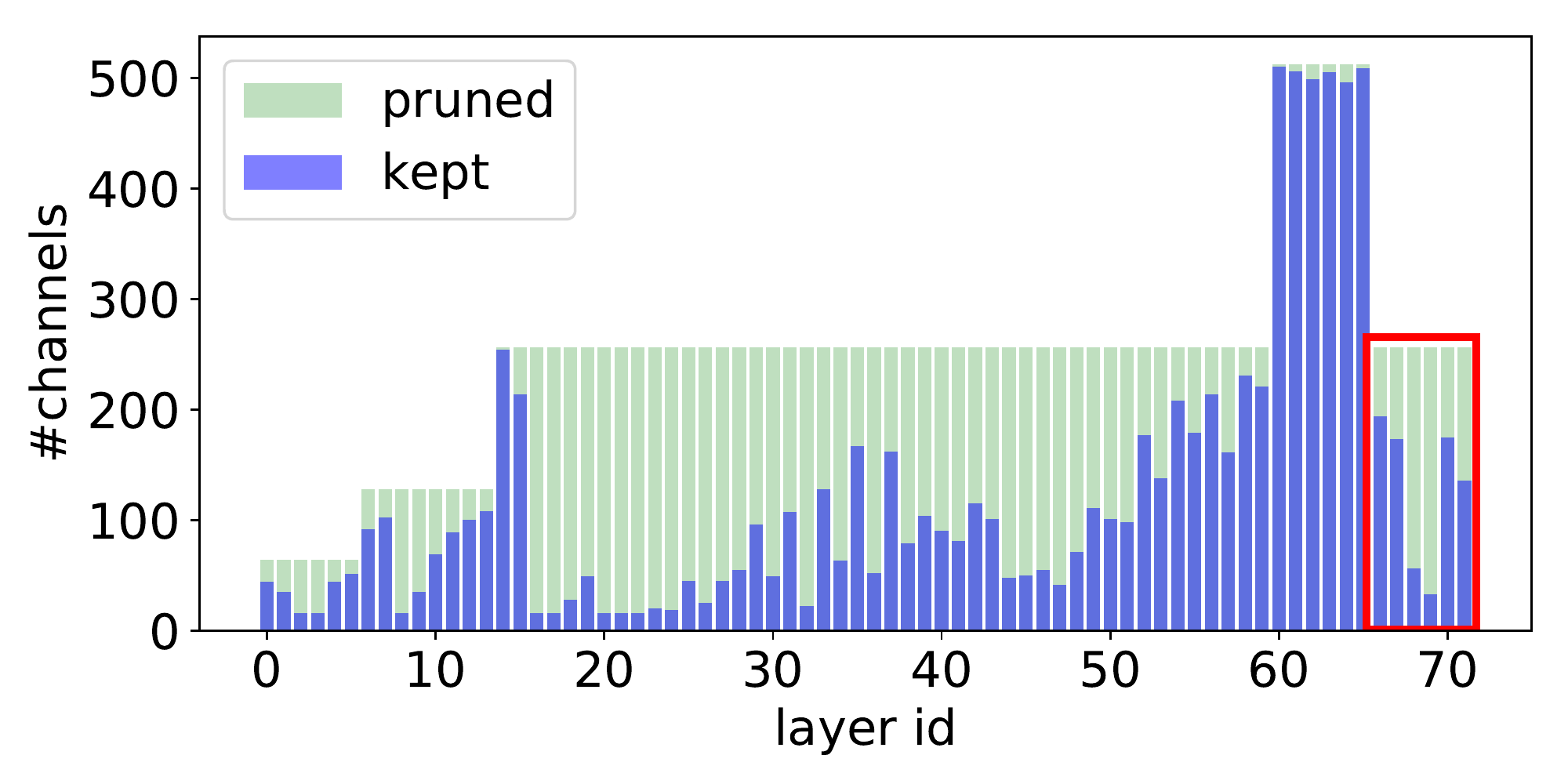}
		\vspace{-1.em}
		\caption{\small Visualization of channels for DeepLabv3 on VOC. }
		\label{fig:vis-cfg-pascal}
	\end{minipage}
	\vspace{-0.9em}
\end{figure}

\subsection{Ablation Studies}
\begin{table}[t]
	\centering
	\small\scalebox{0.95}{
	\begin{tabular}{ll|llll}
		\toprule
		&& mIoU (\%) & \#Para (M) & FLOPs (G)\textsuperscript\textdagger \\ 
		\midrule
		&{DeepLabV3}     & 77.27 & 58.0  & 201.9   \\ 
		\midrule
		\multirow{4}{*}{0.5$\times$} & Slimming-Uni & 73.77 $_{\downarrow 3.50}$ & 36.9 $_{0.64\times}$ & 125.9 $_{0.62\times}$ \\ 
		&Slimming-Ind & 74.91 $_{\downarrow 3.50}$ & 28.6 $_{0.49\times}$ & 101.6 $_{0.50\times}$ \\ 
		&Ours-Uni & 75.33 $_{\downarrow 1.94}$ & 37.6 $_{0.65\times}$ & 131.1 $_{0.65\times}$ \\
		&Ours-Ind & 76.29 $_{\downarrow 0.98}$ & 30.3 $_{0.52\times}$ & 109.9 $_{0.54\times}$  \\ 
		\bottomrule
	\end{tabular}}
	\vspace{-0.5em}
	\caption{\small Ablation results about different pruning strategies on Pascal VOC 2012. The postfix \textit{Ind} means using independent pruning thresholds for backbone and encoder, while \textit{Uni} means using a unified threshold for the whole network. \textsuperscript\textdagger Image size $1080 \times 720$.}
	\label{tab:abla-thres}
\end{table}

{\noindent \bf{Impact of fine-tuning epochs.}} %
Here we explore how the number of fine-tuning epochs on ImageNet and PASCAL VOC 2012 impact the performance of pruned models. As shown in Figure~\ref{fig:fine-tune-epochs}, when we fine-tune the pruned model on PASCAL VOC 2012 for more epochs, the performance improves and becomes stable at around epoch 50. More importantly, the performance is highly competitive even in epoch 20, which demonstrates that the superior performance of our method isn't purely attributed to extra training budgets in fine-tuning. We can also observe that more fine-tuning epochs on ImageNet don't necessarily improve the performance a lot, as shown in Figure~\ref{fig:fine-tune-epochs}. We fine-tuned the pruned models on ImageNet for 20 epochs for a tradeoff between training budget and performance.

{\noindent \bf{Unified vs. Independent pruning thresholds.}} %
As described in Section~\ref{sec:opt}, since the scaling factors in backbone and decoder network are optimized alternately, we instead use two independent pruning thresholds for backbone and decoder of the segmentation network. The ablation results of this strategy are shown in Table~\ref{tab:abla-thres}. These results show that the proposed pruning strategy of using independent thresholds for backbone and decoder boosts the performance of the pruned models of our proposed method and Network Slimming~\cite{slimming}, obtaining higher mIoU with a slightly fewer number of parameters and FLOPs.

{\noindent \bf{Visualization of Pruned Channels.}} %
We visualize the number of channels in each layer for the pruned model obtained by our method on PASCAL VOC 2012 dataset, as shown in Figure~\ref{fig:vis-cfg-pascal}. Channels in red rectangle belong to decoder (ASPP in DeepLabv3) and others belong to backbone (ResNet-101). It can be observed that our method tends to keep more channels in the last residual block of the backbone network. We attribute it to the fact that features in the last block have lower resolution and thus may need more channels to capture the semantic information.

\section{Conclusions}
\label{sec:con}

In this paper we propose a novel multi-task channel pruning method to obtain a lightweight semantic segmentation network.
We first excavate the relationship between the pre-training of the backbone model and the segmentation performance, and then establish an end-to-end multi-task filter pruning approach. The new method simultaneously identifies the redundant filters in both two datasets. Therefore, the produced lightweight segmentation network can greatly maintain the segmentation performance.
Extensive experimental results on several benchmark datasets demonstrate that our method outperforms state-of-the-art pruning methods for generating lightweight segmentation networks.

{\small
\bibliographystyle{IEEEbib}
\bibliography{ref}
}

\clearpage
\section{Appendix}

\subsection{Experimental Settings}

{\noindent \bf{DeepLabv3 on PASCAL VOC 2012.}} %
For training baseline DeepLabv3~\cite{deeplabv3} with ResNet-101 on PASCAL VOC 2012, we use initial learning rate of 0.007 and poly learning rate scheduler where the learning rate is multiplied by $(1-(\frac{iter}{iter_{max}})^{0.9})$. We apply data augmentation including random horizontal flip, random scaling ranging from 0.5 to 2.0 and random crop of $513 \times 513$ during training. We train the model for 50 epochs with batch size of 16 in a Nvidia V100 GPU. 

For multi-task pruning, we set $\alpha_1=0.001$ and $\alpha_2=0.001$. The batch size for ImageNet is 256. The initial learning rate for ImageNet is 0.001 and linear learning rate scheduler is adopted. Since we are using post-activation variant of ResNet, we only prune the first two convolutional layers in each residual block.
For fine-tuning, we set the initial learning rate as 0.0007 and also use poly learning rate policy. The pruned model is fine-tuned on ImageNet for 20 epochs and then fine-tuned on PASCAL VOC 2012 for another 50 epochs.
All inference speeds for different models are evaluated on one Nvidia P100 GPU.

{\noindent \bf{{DeepLabv3 on Cityscapes.}}} %
We use initial learning rate of 0.01 and poly learning rate policy for training baseline DeepLabv3 with ResNet-101 on Cityscapes. We train the model for 480 epochs, with data augmentation of random flip, random scaling (from 0.5 to 2) and random crop with the crop size of $1024 \times 512$. The batch size for training is 32 and 8 Nvidia V100 GPUs are used. 

For multi-task pruning, we set $\alpha_1=0.001$ and $\alpha_2=0.001$. The batch size for ImageNet is 256. The initial learning rate for ImageNet is 0.001 and linear learning rate scheduler is adopted. 
For fine-tuning, we set the initial learning rate as 0.0005 and also use poly learning rate policy. The pruned model is fine-tuned on ImageNet for 20 epochs and then fine-tuned on Cityscapes for another 240 epochs. We employ multi-scale testing of $[0.5, 0.75,1.0,1.25,1.5,1.75]$. When evaluating on \textit{test} set, we train the model on \textit{train\_fine} and \textit{val\_fine} set. No coarse data is used and we do not employ COCO pre-training.

{\noindent \bf{{BiSeNet on Cityscapes.}}} %
BiSeNet-R18~\cite{bisenet}\footnote{ \url{https://github.com/ycszen/TorchSeg}.} is trained with initial learning rate of 0.01, poly learning rate policy and batch size of 32 on Cityscapes for 80 epochs with 8 GPUs. Data augmentations including random flip, random scaling (from 0.75 to 2) and random crop with the crop size of $1536 \times 768$ are utilized.

For multi-task pruning, we set $\alpha_1=0.0001$ and $\alpha_2=0.0001$. The batch size for ImageNet is 512. The initial learning rate for ImageNet is 0.001 and linear learning rate scheduler is adopted. 
For fine-tuning, we set the initial learning rate as 0.001 and also use poly learning rate policy. The pruned model is fine-tuned on ImageNet for 50 epochs and then fine-tuned on Cityscapes for another 80 epochs. Since BiSeNet is devoted for real-time segmentation, no multi-scale testing is adopted in evaluation.

{\noindent \bf{{PSPNet on ADE20K.}}} %
PSPNet-R50~\cite{pspnet} is trained for 120 epochs on ADE20K, with initial learning rate of 0.01 and batch size of 32 on 8 GPUs. Data augmentations of random flip, random scaling (from 0.5 to 2) and random crop with the crop size of $480 \times 480$ are adopted.

For multi-task pruning, we set $\alpha_1=0.0001$ and $\alpha_2=0.0001$. The batch size for ImageNet is 128. The initial learning rate for ImageNet is 0.001 and linear learning rate scheduler is adopted. 
For fine-tuning, we set the initial learning rate as 0.0005 and also use poly learning rate policy. The pruned model is fine-tuned on ImageNet for 50 epochs and then fine-tuned on ADE20K for another 120 epochs. We employ multi-scale testing of $[0.5, 0.75,1.0,1.25,1.5,1.75]$ for evaluation.

\subsection{Detailed Quantitative Results}
\label{sec:intro}
We provide more detailed results including per-category mIoU of different methods on PASCAL VOC 2012 \textit{val} set and Cityscapes \textit{val} set, as shown in Table~\ref{tab:supp-pascal} and Table~\ref{tab:supp-cityscapes} respectively. On PASCAL VOC 2012, the pruned models obtained by our proposed method achieve better mIoU for most categories. On Cityscapes dataset, our method obtains better overall mIoU when the pruning ratio is relatively low (\eg keeping about 75\% of channels).

\begin{table*}[htb]
	\centering
	\setlength{\tabcolsep}{1.1mm}
	\small\scalebox{0.75}{
		\begin{tabular}{l|*{21}{c}|c}
			\toprule
			Model & bg & aero & bike & bird & boat & bottle & bus & car & cat & chair & cow & table & dog & horse & mbike & person & plant & sheep & sofa & train & tv & mean  \\
			\midrule
			DeepLabv3~\cite{deeplabv3} & 94.05 & 87.41 & 41.42 & 88.48 & 71.23 & 82.25 & 93.9 & 88.36 & 92.86 & 40.08 & 85.93 & 53.75 & 89.1 & 84.69 & 84.25 & 85.09 & 60.68 & 87.93 & 49.15 & 85.32 & 76.82 & 77.27 \\ 
			\midrule
			Uniform 0.75 & 93.50 & 89.66 & 41.35 & 87.12 & 71.05 & 78.43 & 93.51 & 84.04 & 90.56 & 38.22 & 83.07 & 47.84 & 85.75 & 81.37 & 78.9 & 83.89 & \bf 66.29 & 80.54 & 43.5 & \bf 84.16 & 74.14 & 75.09 \\ 
			Slimming 0.75$\times$~\cite{slimming} & 93.82 & 89.68 & 41.71 & 88.91 & \bf 73.68 & 80.29 & 92.37 & \bf 87.95 & 92.51 & 41.53 & 85.63 & \bf 49.64 & 87.13 & 84.7 & 83.52 & \bf 85.21 & 59.41 & \bf 87.95 & 47.41 & 81.23 & 75.07 & 76.64 \\ 
			MTP 0.75$\times$ (Ours) & \bf 93.91 & \bf 88.92 & \bf 42.13 & \bf 90.70 & 73.28 & \bf 80.69 & \bf 93.62 & 87.61 & \bf 93.46 &\bf  44.07 & \bf 86.48 & 46.81 & \bf 89.31 & \bf 87.96 & \bf 84.01 & 85.38 & 61.41 & 84.09 & \bf 48.56 & 83.04 & \bf 77.33 & \bf 77.28 \\ 
			\midrule
			ThiNet 0.5$\times$~\cite{thinet} & 93.57 & 87.51 & 40.47 & \bf 88.19 & \bf 70.02 & 74.85 & 93.32 & 85.85 & \bf 92.79 & 37.05 & 86.13 & 48.59 & \bf 87.07 & 84.02 & 82.86 & 83.66 & 52.22 & 81.10 & 44.02 & 82.76 & 72.87 & 74.71 \\ 
			Slimming 0.5$\times$~\cite{slimming} & 93.35 & \bf 88.94 & 40.61 & 86.46 & 69.15 & 76.98 & 91.89 & 86.35 & 92.26 & 38.21 & 84.86 & 47.72 & 86.29 & 84.69 & \bf 83.58 & 83.71 & 53.77 & 84.64 & 44.8 & 81.42 & 73.38 & 74.91 \\ 
			MTP 0.5$\times$ (Ours) & \bf 93.68 & 88.45 & \bf 41.01 & 86.93 & 69.2 & \bf 77.61 & \bf 93.61 & \bf 87.07 & 91.57 & \bf 39.00 & \bf 88.63 & \bf 49.10 & 86.76 & \bf 86.62 & 82.31 & \bf 84.37 & \bf 60.42 & \bf 86.66 & \bf 48.63 & \bf 85.58 & \bf 74.78 & \bf 76.29 \\ 
			\bottomrule
		\end{tabular}
	}
	\caption{Detailed results of different methods on PASCAL VOC 2012 \textit{val} set.}
	\label{tab:supp-pascal}
\end{table*}

\begin{table*}[htb]
	\centering
	\setlength{\tabcolsep}{1.1mm}
	\small\scalebox{0.8}{
		\begin{tabular}{l|*{19}{c}|c}
			\toprule
			& road & sidewalk & bld & wall & fence & pole & light & sign & vgttn & terrain & sky & person & rider & car & truck & bus & train & mbike & bike & mean \\
			\midrule
			DeepLabv3~\cite{deeplabv3} & 98.06 & 84.32 & 92.47 & 59.81 & 61.94 & 58.55 & 69.68 & 78.13 & 92.30 & 64.68 & 94.18 & 81.09 & 64.07 & 94.95 & 81.74 & 89.66 & 81.47 & 70.13 & 77.20 & 78.65 \\
			\midrule
			Slimming 0.75$\times$~\cite{slimming} & 98.19 & 85.33 & 92.40 & 60.67 & \textbf{63.03} & \textbf{57.28} & \textbf{69.24} & 77.05 & \textbf{92.27} & \textbf{67.79} & \textbf{94.11} & 80.62 & \textbf{63.96} & 94.70 & \textbf{78.97} & 89.93 & 77.07 & 69.56 & 76.87 & 78.37 \\
			MTP 0.75$\times$ (Ours) & \textbf{98.27} & \textbf{85.69} & \textbf{92.45} & \textbf{62.76} & 62.57 & 55.68 & 68.34 & \textbf{77.18} & 92.12 & 65.58 & 94.10 & \textbf{80.67} & 63.24 & \textbf{94.82} & 78.84 & \textbf{91.24} & \textbf{82.34} & \textbf{70.51} & \textbf{77.05} & \textbf{78.60} \\
			\midrule
			Slimming 0.5$\times$~\cite{slimming} & 97.95 & 83.78 & \textbf{92.00} & 59.59 & \textbf{62.00} & 54.17 & 66.50 & 74.87 & 91.87 & 64.18 & 93.46 & 78.72 & 59.93 & 94.23 & 73.51 & \textbf{89.55} & \textbf{82.72} & 67.38 & 75.51 & 76.94 \\
			MTP 0.5$\times$ (Ours) & \textbf{98.15} & \textbf{84.62} & 92.14 & \textbf{61.30} & 61.69 & \textbf{55.05} & \textbf{67.55} & \textbf{76.18} & \textbf{91.97} & \textbf{64.86} & \textbf{93.89} & \textbf{79.93} & \textbf{62.20} & \textbf{94.60} & \textbf{78.17} & 87.76 & 75.95 & \textbf{68.49} & \textbf{76.00} & \textbf{77.39}\\ 
			\bottomrule
		\end{tabular}
	}
	\caption{Detailed results of different methods on Cityscapes \textit{val} set.}
	\label{tab:supp-cityscapes}
\end{table*}

\subsection{Qualitative results}

\begin{figure*}[t]
	\footnotesize
	\begin{center}
		\begin{tabular}{c}
			\includegraphics[width = 0.85\linewidth]{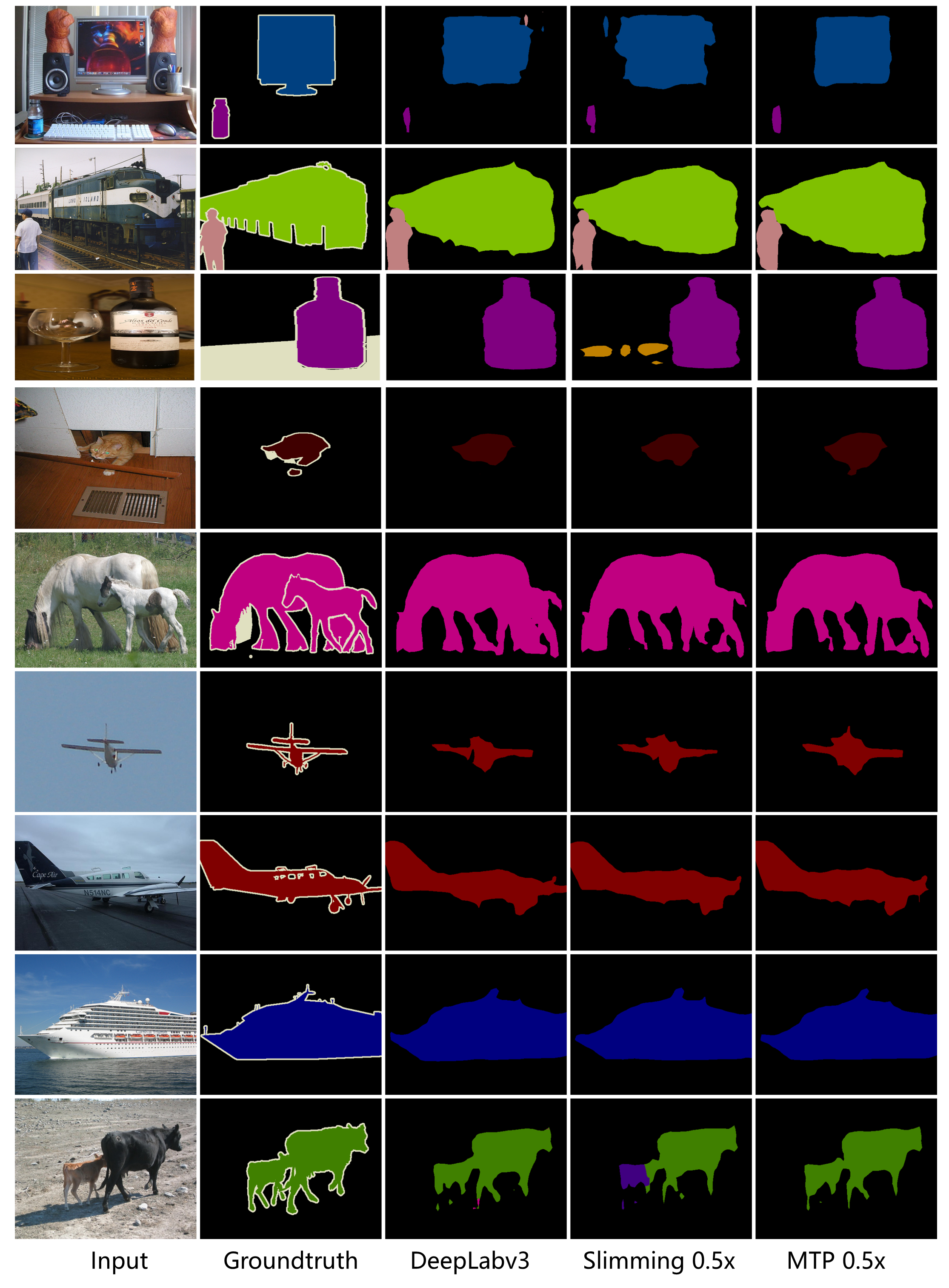} \\
		\end{tabular}
	\end{center}
	\caption{Qualitative results of different methods on PASCAL VOC 2012 \textit{val} set. From left to right are input images, ground truth, results of baseline DeepLabv3, Network Slimming 0.5$\times$ and Our method 0.5$\times$ respectively.}
\label{fig:vis_sup_pascal}
\end{figure*}

\begin{figure*}[t]
\footnotesize
\begin{center}
	\begin{tabular}{c}
		\includegraphics[width = 0.85\linewidth]{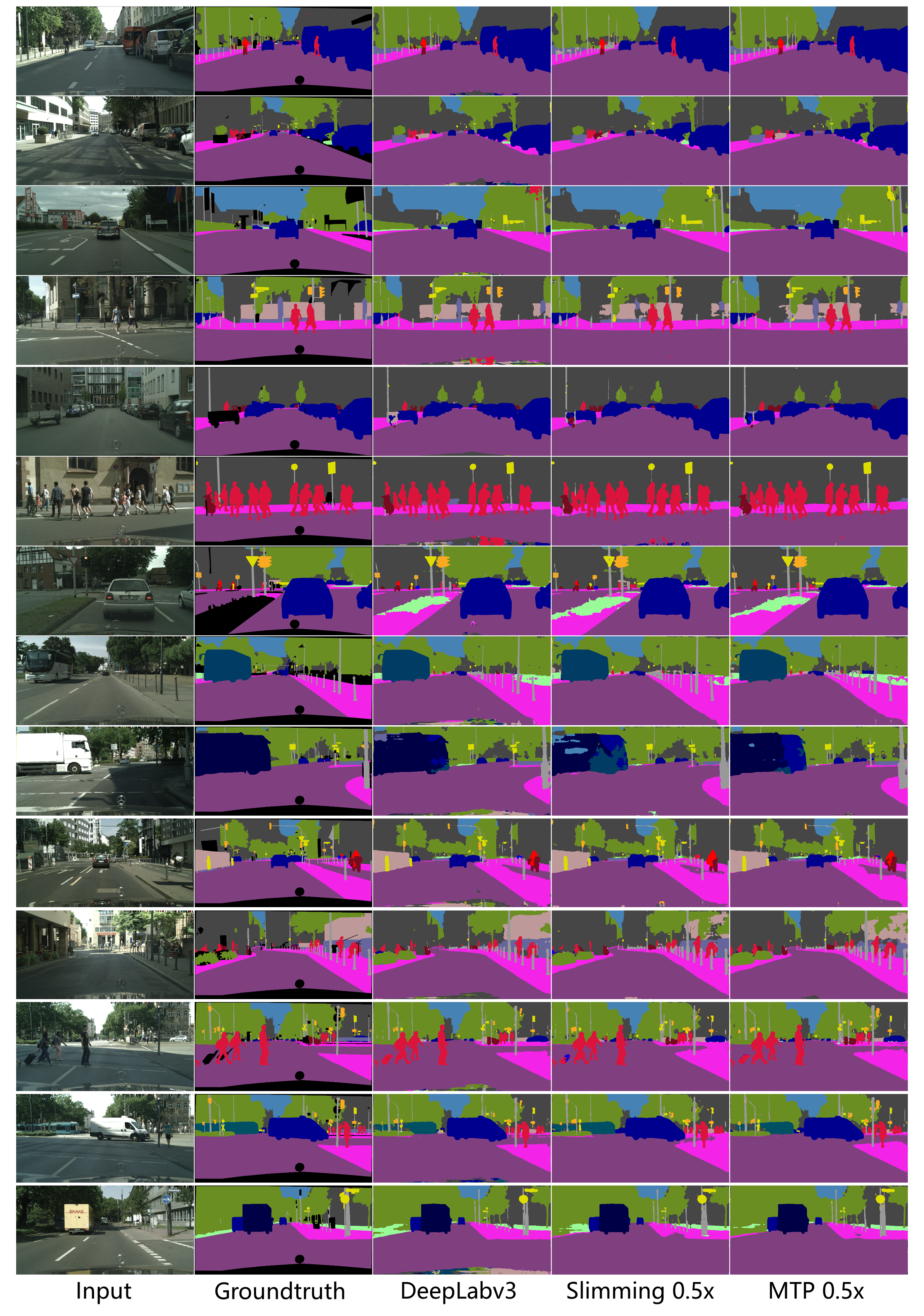} \\
	\end{tabular}
\end{center}
\caption{Qualitative results of different methods on Cityscapes \textit{val} set.  From left to right are input images, ground truth, results of baseline DeepLabv3, Network Slimming 0.5$\times$ and Our method 0.5$\times$ respectively.}
\label{fig:vis_sup_cityscapes}
\end{figure*}

\begin{figure*}[t]
\footnotesize
\begin{center}
	\begin{tabular}{c}
		\includegraphics[width = 0.85\linewidth]{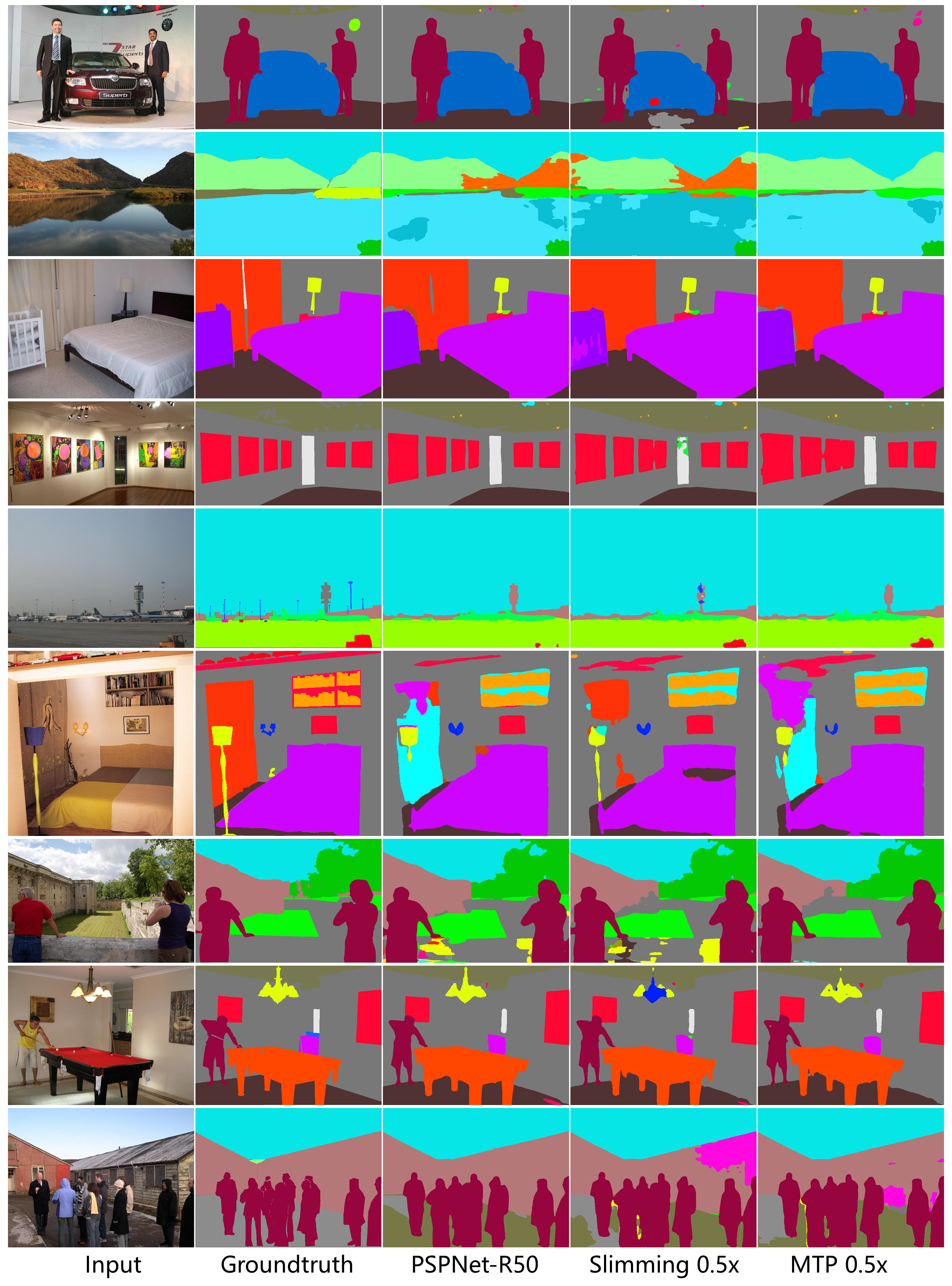} \\
	\end{tabular}
\end{center}
\caption{Qualitative results of different methods on ADE20K \textit{val} set.  From left to right are input images, ground truth, results of baseline PSPNet-R50, Slimming 0.5$\times$ and Our method 0.5$\times$ respectively.}
\label{fig:vis_sup_ade}
\end{figure*}

Here we provide more qualitative results of baseline DeepLabv3~\cite{deeplabv3}, Network Slimming 0.5$\times$~\cite{slimming} and Our method 0.5$\times$ on PASCAL VOC 2012 \textit{val} set, as shown in Figure~\ref{fig:vis_sup_pascal}. Note that our proposed method obtains better segmentation results than Network Slimming~\cite{slimming}. More qualitative results on Cityscapes \textit{val} set are shown in Figure~\ref{fig:vis_sup_cityscapes}.
Qualitative results of baseline PSPNet-R50~\cite{pspnet}, Slimming 0.5$\times$ and MTP 0.5$\times$ are shown in Figure~\ref{fig:vis_sup_ade}.

\begin{figure}[h]
	\centering
	\includegraphics[width = 0.99\linewidth]{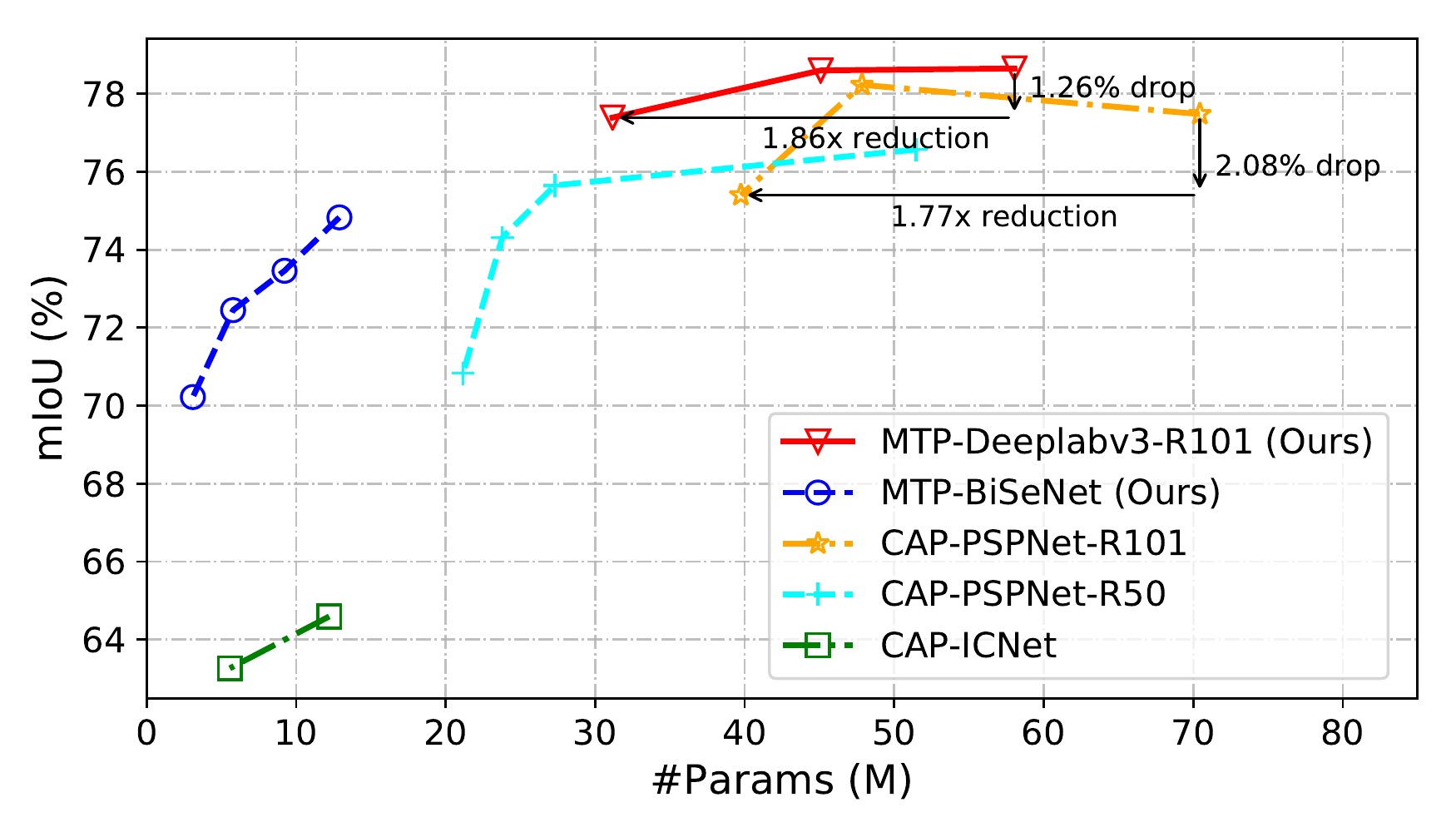}
	\vspace{-1.5em}
	\caption{\small {Comparisons of pruned models on Cityscapes val set.}}
	\label{fig:comp_cap}
\end{figure}

\begin{table}[htb]
	\centering
	\small\scalebox{0.75}{
	\begin{tabular}{c|ccc|cc}
			\toprule
			\multirow{2}{*}{Acc (\%)} & &{0.75$\times$}& & {0.5$\times$}&\\
			& Uniform  & Slimming~\cite{slimming} & \bf MTP (Ours) & Slimming ~\cite{slimming} &\bf  MTP(Ours)      \\
			\midrule
			Top1 & 75.51     & 75.04   & \bf 76.03     & 71.07    & \bf 73.15         \\
			Top5 & 92.55     & 92.44  & \bf 92.98     & 90.24   & \bf 91.50        \\
			\bottomrule
		\end{tabular}
	}
	\caption{Evaluation of the pruned backbone networks on ImageNet~\textit{val} set.}
	\label{tab:imagenet}
\end{table}

\begin{figure}[tb]
	\centering
	\includegraphics[width = 0.9\linewidth]{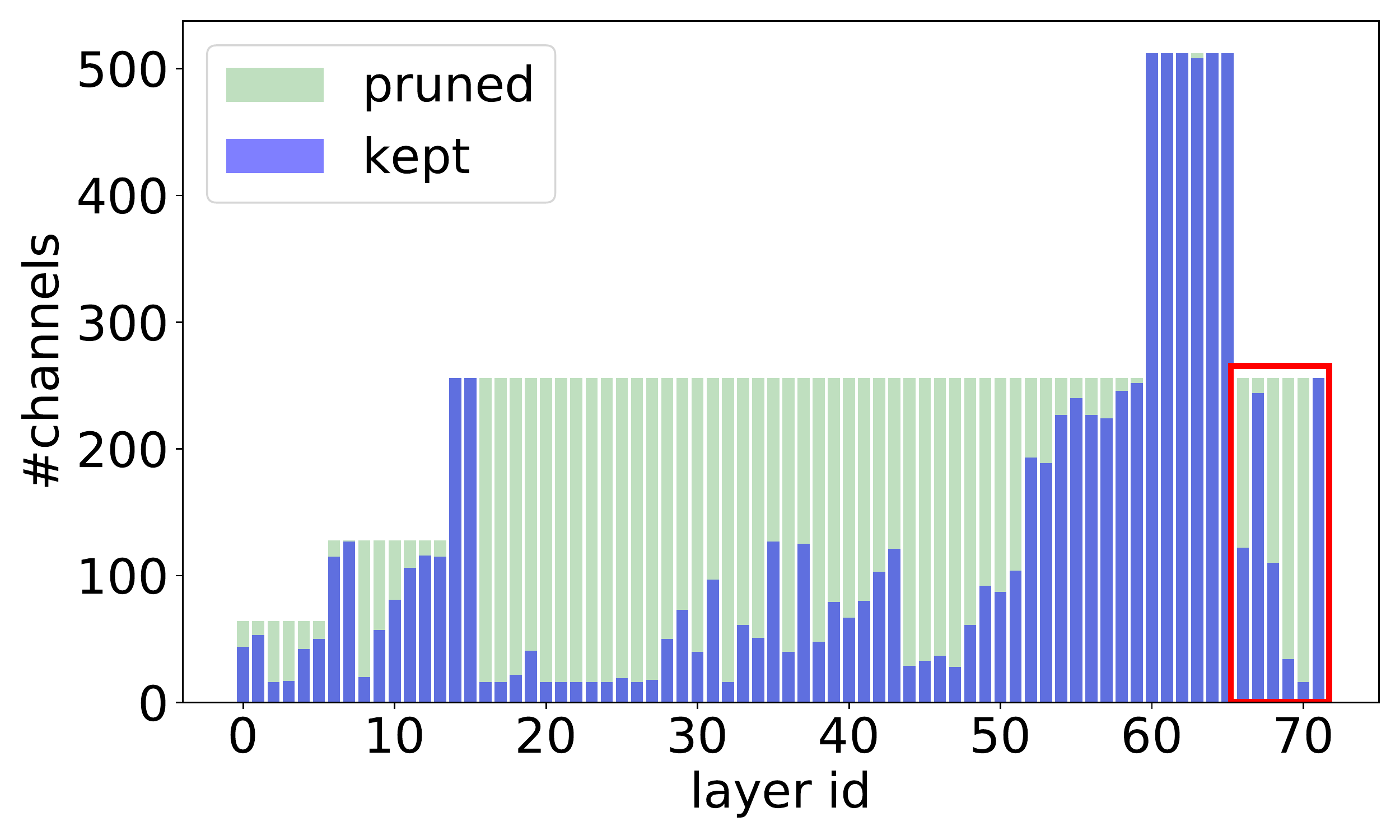}
	\caption{Visualization of pruned channels for DeepLabv3 on Cityscapes.}
	\label{fig:vis-cfg-cityscapes}
\end{figure}

\subsection{Additional Comparisons with state-of-the-arts}
\label{sec:intro}

{We also compare our proposed method with CAP~\cite{He_2021_WACV} on Cityscapes val set. As shown in Figure~\ref{fig:comp_cap}, our method only suffers from 1.26\% mIoU drop for DeepLabv3 when reducing 44\% FLOPs while CAP has 2.08\% mIoU drop for PSPNet on Cityscapes.}

\subsection{Performance on Classification Task.} %
Since the proposed method simultaneously discovers the pruned architecture on classification and segmentation tasks, it is expected that our pruned models also have a good performance on image classification. We evaluate the pruned backbone network of DeepLabv3 on ImageNet \textit{val} set. As shown in Table~\ref{tab:imagenet}, the pruned models obtained by our proposed method consistently outperform Network Slimming~\cite{slimming} and uniformly pruned baseline at different pruning ratios, which demonstrates the advantage of the proposed multi-task pruning scheme.

\subsection{Visualization of Pruned Channels on Cityscapes.} %
We visualize the pruned channels for Deeplabv3 on Cityscapes, as shown in Figure~\ref{fig:vis-cfg-cityscapes}. The observations for pruned channels for PASCAL VOC 2012 still hold for Cityscapes. More interesting, we find that different datasets prefer different configuration of decoder (ASPP). For example, pruned model for Cityscapes has much fewer channels in image pooling branch of ASPP (see the $70^{th}$ channel) than PASCAL VOC, which may indicate that global context information is more essential for more complex scenes.

\end{document}